\documentclass{article}

\usepackage{graphicx}
\usepackage{amsmath,amsfonts,amssymb,amscd,amsthm,xspace}
\usepackage{natbib}            
\usepackage{mathtools}
\usepackage{algorithm}
\usepackage{algorithmic}
\usepackage{threeparttable}
\usepackage{multirow}
\usepackage{algorithmic}
\usepackage{epstopdf}
\usepackage[scriptsize]{subfigure}
\usepackage{booktabs}
\usepackage{color}

\title{Parameter Estimation in Computational Biology by Approximate
Bayesian Computation coupled with Sensitivity Analysis}
\author{Xin Liu and Mahesan Niranjan}
\date{\today}

\begin{document}


\maketitle

\abstract{\textbf{Motivation:}
We address the problem of parameter estimation in models of systems biology from noisy observations. The models we consider are characterized by simultaneous deterministic nonlinear differential equations whose parameters are either taken from {\em in vitro} experiments, or are hand-tuned during the model development process to reproduces observations from the system. We consider the family of algorithms coming under the Bayesian formulation of Approximate Bayesian Computation (ABC), and show that sensitivity analysis could be deployed to quantify the relative roles of different parameters in the system. Parameters to which a system is relatively less sensitive (known as sloppy parameters) need not be estimated to high precision, while the values of parameters that are more critical (stiff parameters) need to be determined with care. A tradeoff between computational complexity and the accuracy with which the posterior distribution may be probed is an important characteristic of this class of algorithms. \\
\textbf{Results:}
With sensitivity analysis identifying the relative roles of different parameters, we show that a judicious allocation of computational budget can be made, achieving efficient parameter estimation. We demonstrate a three stage strategy, in which after identifying different parameters as sloppy and stiff, a coarse, computationally inexpensive step could be used to set the sloppy parameters. This is followed by an optimization stage with more stringent sample acceptance criteria to estimate stiff parameters. We also apply a recently developed adaptive algorithm for gradual cooling of the acceptance threshold within the ABC framework. We demonstrate the effectiveness of the proposed method on three models of oscillatory systems and one of transient response of an organism to heat shock, taken from the systems biology literature.\\
\textbf{Contact:}{M.Niranjan@Southampton.ac.uk}
}

\section{Introduction}
Computational modeling of biological systems is about describing quantitative relationships of biochemical reactions by systems of differential equations \citep{Kitano01032002}. Knowledge of biological processes captured in such equations, when solutions to them match measurements made from the system of interest, help confirm our understanding of systems level function. Examples of such models include cell cycle progression \citep{chenCellCycle}, integrate and fire generation of heart pacemaker pulses \citep{zhangPace2000} and cellular behavior in synchrony with the circadian cycle \citep{Leloup10062003}. A particular appeal of modeling is that models can be interrogated with {\em what if} type questions to improve our understanding of the system, or be used to make quantitative predictions in domains in which measurements are unavailable.

A central issue in developing computational models of biological systems is setting parameters such as rate constants of biochemical reactions, synthesis and decay rates of macromolecules, delays incurred in transcription of genes and translation of proteins, and sharpness of nonlinear effects (Hill coefficient) are examples of such parameters. Parameter values are usually determined by conducting {\em in vitro} experiments ({\em e.g.} \citep{Niedenthal1996, Wadsworth2001, Tseng2002, Wiedenmann2004}). When parameter values are not available from experimental measurements, modelers often resort to hand-tuning during the model development process and publish the range of values of a parameter required to achieve a match between model output and observed data. In dynamical systems characterized by variations over time, concentrations of different molecular species (proteins, metabolites etc.) may also be of interest. In this setting, we encounter two difficulties. First, parameters measured by {in vitro} experiments may not be good reflections of the {\em in vivo} reality. And, second, some parameters in a system may not be amenable to experimental measurements. These limitations motivate the need to infer parameters in a computational model based on input-output observations and recent literature on computational and systems biology has seen intense activity along these lines \citep{Barenco2006, Vyshemirsky2008, Jayawardhana2008, wilkinsonHandbook2010, Xin2012}.

One way of setting parameters systematically is based in techniques for search and optimization. For example, \cite{Mendes1998} compared several optimization based algorithms for estimating parameters along biochemical pathways, concluding that no single approach significantly outperforms other available approaches. Similar work on a developmental gene regulation circuit is described in \cite{FomekongGapGene2009}, and on a spline approximation based method for learning the parameters of enzyme kinetic model in a cell cycle system is described in \cite{zhan2011}.

An alternate approach is the use of probabilistic Bayesian formulations to quantify uncertainties in the process of estimating parameters. Work described in \cite{golightly2005, dewar2010, Barenco2006, Vyshemirsky2008, Jayawardhana2008} fall into this category. With time varying or dynamical systems, some authors have pointed out advantages of sequential estimation models, formulating the problem as state and parameter estimation in a state-space modeling framework \citep{quach2007, sun2008, Gabriele2010}. Kalman filtering and its variants, and nonparametric particle filtering have been applied, in this setting \citep{yang2007, Xin2012}.

An approach that has attracted much interest recently is the method of Approximate Bayesian Computation (ABC) or likelihood-free inference. While this approach has its roots in population genetics \citep{Tavare1997}, where the likelihood is usually too complicated to write down in a computable form, it is also seen as a viable tool in systems biology parameter estimation problems (\citep{Tina2009, Maria2009}). In brief, the ABC approach assumes it is easier to simulate data from a model of interest than it is to compute and work with its likelihood under some assumed noise model. Hence a structured search could be carried out, in which one repeatedly simulates with parameter values sampled from a prior distribution, computes the error between simulated and observed data and decides to retain or reject a sample based on this error exceeding a threshold. Samples thus retained define a posterior density in the space of parameters to be estimated.

Several enhancements to this basic scheme of \cite{Tavare1997} have been developed. \cite{Beaumont2002} used a weighted linear regression adjustment to improve the efficiency of search (ABC-Regression) with interesting applications \citep{Hamilton2005}. In systems biology, \cite{Tina2009} proposed a Sequential Importance Sampling (SIS) based ABC approach to infer the parameters of Lotka-Volterra and repressilator models. This ABC-SIS method was also utilized by \cite{Maria2009} to estimate the parameters of mitogen-activated protein kinase (MAPK) signaling pathway \citep{Krauss2008}.

In implementing ABC algorithms, an inherent compromise between computational efficiency and accuracy of inference has to be made. This compromise is in the form of the acceptance threshold ($\epsilon$), also referred to as tolerance. Setting $\epsilon$ to a high value results in a large number of acceptances but the resulting inference will be inaccurate. Researchers use arbitrary user-specified approaches \citep{Sisson2007, Robert2009,Tina2009} to deal with this issue, i.e. start with a large $\epsilon$ and progressively reduce in value as the algorithm converges. In an elegant piece of work recently, \cite{DelMoral2012} showed how such an $\epsilon$ scheduling approach can be adaptively set. For the ABC class of algorithms this adaptive technique is indeed a major breakthrough.

\cite{Chiachio2014} demonstrate the effectiveness of this adaptive ABC method by estimating parameters of two stochastic benchmark models (the second order moving average process and the white noise corrupt linear oscillator).
\cite{Hainy2016} also used this algorithm to learn parameters in the spatial extremes model from the data on maximum annual summer temperatures from 39 sites in the Midwest region of the USA.
In epidemiology, \cite{Lintusaari2016} utilized such likelihood-free approach to identify of transmission dynamic models for infectious diseases.
\color{black}

An additional point to consider, which we pursue in this work, is that in models of Systems Biology not all parameters behave in the same way. \cite{Gutenkunst2007} observe that in a large number of models, systems level behavior is not sensitive to some parameters while being critically sensitive to others. These are referred to as `sloppy' and `stiff' parameters, respectively. Over a wide range of values with sloppy parameters, system output varies very slightly, while even small changes to the setting of stiff parameters cause large perturbations in outputs. The ABC framework, with its analysis by synthesis structure, is ideally suited to exploit this observation as we demonstrate in this work.
Additionally, we comment on the inherent compromise between accuracy and computational efficiency in choosing the number of particles and parallel filters for the adaptive ABC-SMC algorithm.

\begin{figure}[t]
\centering
\includegraphics[width=0.30\textwidth]{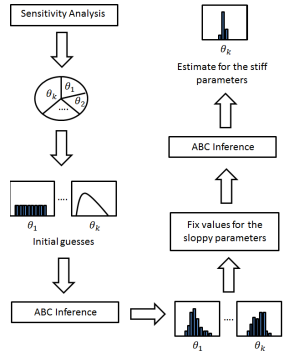}
\caption{Computational steps in the proposed approach: Carrying the global sensitivity analysis technique (extended Fourier amplitude sensitivity test), we identify the sloppy and stiff parameters in system. Following this, starting from an initial guess of parameter values (most likely to be non-informative), we carry out estimation of parameters by Approximate Bayesian Computation (ABC) method associated with a coarse acceptance criterion. In the examples considered, we are using the advanced ABC algorithm for having the tolerance schedule adaptively. Details are described in \textbf{Section} \ref{ABCSMC}. The sloppy parameters are then fixed to values determined by this coarse analysis. In the final stage, we estimate the stiff parameters of the system by running the ABC method to tighter error tolerance. This achieves a selective partitioning of the computational budget, and reliable estimates can be achieved within reasonable times.
}
\label{Figure:ABCSA_Diagram}
\end{figure}

\section{Methods}
The computational approach we propose, shown schematically in Fig. \ref{Figure:ABCSA_Diagram}, is a combination of sensitivity analysis \citep{Saltelli1999} and advanced ABC-SMC \citep{DelMoral2012}. We choose a particular method of sensitivity analysis (global sensitivity analysis, discussed later), to group the unknown parameters of a model into stiff and sloppy parameters. The sloppy parameters are estimated by a coarse search, which in ABC algorithms is achieved by using large values for acceptance thresholds. Once these are set, stiff parameters are estimated with adaptive evaluation of acceptance thresholds over a finer range.

We now summarize the methods of ABC-SMC with adaptive acceptance tolerance and the global sensitivity analysis techniques used in this work. We start with a succinct, tutorial type description of the ABC-SMC and Sensitivity Analysis frameworks, and present implemetation details to \textbf{Supplementary Material}.

\subsection{Adaptive ABC-SMC}\label{ABCSMC}
In parameter estimation problems, we seek a posterior distribution over parameters given observations, denoted $p(\boldsymbol{\theta}|\textbf{X})$. As noted in the Introduction, the basic idea in ABC algorithms is to sample the unknown from a prior distribution, $\theta^{\ast}\sim\pi(\theta)$, synthesize data from the model under study, $\textbf{X}^{\ast} \sim f(\textbf{x}_{0}, \theta^{\ast})$, where $\textbf{x}_{0}$ is the initial condition and $f(\cdot,\cdot)$ is the model, and accept $\theta^{\ast}$ as a sample for the posterior if the synthesized data $\textbf{X}^{\ast}$ (referred to pseudo-observations in the following descriptions) is close enough in some sense to the observations $\textbf{X}$. In its earliest form \citep{Tavare1997}, the generated particle $\theta^{\ast}$ was accepted only if $\textbf{X}^{\ast}$ was identical to the observations $\textbf{X}$. It became immediately evident that this is an inefficient procedure because thousands of trials needed to be performed before accepting one of the generated particles. A modification to the scheme, introduced by \cite{Pitt1999} was to define a tolerance $\epsilon$ and accept particles when the discrepancy between $\textbf{X}^{\ast}$ and $\textbf{X}$ was smaller than this. In our discussion with Systems Biology models, we will focus on $f(\cdot,\cdot)$ being a set of ordinary differential equations which can be numerically integrated, and use Euclidean distance between the pseudo-observations and the model output as measure of discrepancy.

The Sequential Monte Carlo (SMC) method has recently been applied within the ABC setting \citep{Sisson2007, Tina2009, Robert2009}, in which, instead of synthesis from a single sample drawn from the prior being tested for acceptance of the sample, a population of samples is drawn, and progressively perturbed and synthesis from them tested for acceptance. At any stage, particles are drawn by sampling from the population of the previous stage in proportion to weights associated with them and perturbed via a transition kernel ${\theta}^{\ast\ast}\sim k({\theta}^{\ast})$. Pseudo-observations are then synthesized from the underlying model for the new population of samples: $\textbf{X}^{\ast}\sim f(\textbf{x}_{0}, {\theta}^{\ast\ast})$. Particle ${\theta}^{\ast\ast}$ is accepted and weighted if the discrepancy between synthetic data $\textbf{X}^{\ast}$ and true dataset $\textbf{X}$ is lower than the current tolerance $\epsilon_{t}$. To preserve diversity of particles, {\em i.e.} when \{$\sum_{i=1}^{N_{\text{smc}}}{(w_{t}^{i})^{2}}\}^{-1}\leq \frac{N_{\text{smc}}}{2}$, resampling is also applied \citep{doucetBook}. This population based approach is shown to result in more effective exploration of the space.
Additionally, as mentioned before, the acceptance criterion is gradually reduced sequence of tolerance thresholds $\boldsymbol{\epsilon}=\{\epsilon_{1},\ldots,\epsilon_{T}\}$ at \cite{Sisson2007,Robert2009,Tina2009}, in an arbitrary, user-defined way.

\cite{DelMoral2006} introduced a SMC sampler which
constructs $M_{\text{smc}}$ (referred to integer factor) Markov kernels
being operated in parallel. In the series of ABC developments,
\cite{DelMoral2012} extended this concept to the ABC setting
and formed an adaptive algorithm to determine the tolerance
schedule $\boldsymbol{\epsilon}$.
The weight calculation of this SMC sampler is given as
\begin{align}\label{weightABCSMC}
{w}_{t}^{i}&\propto {w}_{t-1}^{i} \frac{\sum_{m=1}^{M_{\text{smc}}}{\mathcal{I}_{\text{A}_{\epsilon_{t}}}(\textbf{S}_{m,t-1}^{\ast,i},\textbf{X})}}
{\sum_{m=1}^{M_{\text{smc}}}{\mathcal{I}_{\text{A}_{\epsilon_{t-1}}}(\textbf{S}_{m,t-1}^{\ast, i},\textbf{X})}}
\end{align}
where $\textbf{S}^{\ast} \in \mathbb{R}^{M_{\text{smc}} \times N_{\text{smc}} \times D_{\text{s}} \times N_{\text{OT}}}$ are the pseudo-observations,
and $\textbf{S}_{m,t-1}^{\ast, i}$ is the sub-block of pseudo-observations ($\textbf{S}_{m,t-1}^{\ast, i}$ is equivalent to aforementioned $\textbf{X}^{\ast}$) can be interpreted as the $m^{th}$ synthetic
outputs generated by the $i^{th}$ parameter particle at $t$ iteration $\boldsymbol{\theta}_{t-1}^{i}$.
$D_{\text{s}}$ and $N_{\text{OT}}$ are the dimension of states in system and
the number of data points in output, respectively.
$\mathcal{I}_{\epsilon_{t}}(\textbf{S}_{m,t-1}^{\ast, i},\textbf{X})$
is an indicator function that returns one if the discrepancy between
pseudo-observation $\textbf{S}_{m,t-1}^{\ast i}$ and data $\textbf{X}$
is less than the tolerance $\epsilon_{t}$, zero otherwise.
$\text{A}$ is the discrepancy, and which is measured by an Euclidean
distance in this work.

By taking advantage of distribution of particles in the parallel filters,
ABC-SMC is able to adaptively select the current
tolerance level $\epsilon_{t}$.
The idea behind this automatic scheme
is to determine an appropriate reduction of the tolerance level
based on the proportion of particles
surviving under the current tolerance. If a large amount
of particles remain `alive', it implies the
acceptance criterion is relatively loose and it is
safe to make a jump for the next tolerance
level. In contrast, if the ratio of `alive' particles
is low, this means that particles are less likely to describe
the posterior and therefore, a tiny movement should be considered.

Given the proportion of `alive' particles by
\begin{align}\label{SMCPA}
\text{PA}(\textbf{w}_{t},\epsilon_{t})=\frac{\sum_{i=1}^{N_{\text{smc}}}{\mathbb{I}_{(0,\infty)}({w}_{t}^{i})}}{N_{\text{smc}}}
\end{align}
the next tolerance level $\epsilon_{t+1}$ is chosen by making such proportion equals to a given percentage of the current tolerance
$\text{PA}(\textbf{w}_{t},\epsilon_{t+1})
=\alpha_{\text{smc}} \text{PA}(\textbf{w}_{t},\epsilon_{t})$,
where $\alpha_{\text{smc}} $ is the tolerance reduction factor.
The search of $\epsilon_{t+1}$ is achieved by {\tt fzero} function in Matlab via setting the starting point to $\epsilon_{t}$.
\color{black}

To solve ODEs in use of running filters parallel, the initial conditions are set to $\widehat{\textbf{X}}_{0} \sim \mathcal{U}(\textbf{x}_{0},k\textbf{x}_{0})$, where $\widehat{\textbf{X}}_{0} \in \mathbb{R}^{M_{\text{smc}} \times \text{D}_{\text{s}}}$, $\textbf{x}_{0}$ is `true' initial values for generating observations and $k$ is the scaling factor to control consistency/inconsistency of initial conditions across filters.
\color{black}
Additionally, for moving particles to explore the parameter space,
similarly to the previous work \citep{Xin2012}, we use the kernel smoothing
with shrinkage as parameter evolution \citep{Liu2001}, instead of random walk kernel.
The details and pseudo-code of ABC-SMC can be found in
\textbf{Section 2} of \textbf{Supplementary materials}
accompanying the online version of this paper.

The Matlab code of proposed method is available at:
https://github.com/brianliu2/ABCSMC-SA
\color{black}

\subsection{Extended Fourier Amplitude Sensitivity Test}
The {\em extended Fourier amplitude sensitivity test} (eFAST) \citep{Saltelli1999}
is one of the global sensitivity analysis techniques, which is derived by decomposing
variance and claimed to be applicable for analyzing the nonlinear and non-monotonic systems.
It has been successfully deployed to analyze the property of parameters in
TCR-activated MAPK signalling pathway model \citep{Zheng2006} and thermodynamics gene expression model \citep{Dresch2010}.
The algorithm initially partitions the total
variance of the data, evaluating what fraction
of the variance can be determined by variations in the parameter of interest.
This quantity, known as the sensitivity index,
defines the property of stiff or sloppy, is calculated as
\begin{align}\label{sensitivity}
\text{SI}=\frac{\text{Var}_{\boldsymbol{\theta}}[\text{E}(\textbf{X}^{\ast}|\boldsymbol{\theta})]}
{\text{Var}(\textbf{X}^{\ast})}=\frac{V_{\theta}}{V}.
\end{align}
The variance is calculated with respect to sets of synthesis
which are generated by solving the model associated with samples of
parameters.
In order to draw samples for parameters,
\cite{Saltelli2002} derived a heuristic sinusoidal function, given as
\begin{align}\label{Gfunction}
\boldsymbol{\theta}&=G(\sin(\boldsymbol{\omega}\boldsymbol{s}+\boldsymbol{\varphi}))\nonumber \\
&=\frac{1}{2}+\frac{1}{\pi}\arcsin(\sin(\boldsymbol{\omega}\boldsymbol{s}+\boldsymbol{\varphi})),
\end{align}
where $\boldsymbol{\varphi}$ is the $D_{\text{p}}\times N_{\text{se}}$ matrix for
providing random phase-shifting and which follows uniformly distributed $[0,2\pi]$.
$N_{\text{se}} $ specifies the number of samples drawn from
the function $G(\cdot)$. $D_{\text{p}}$ is the dimension of parameters in system.
$\boldsymbol{s}$ in Eqn. (\ref{Gfunction}) defines a $1\times N_{\text{se}} $ equally spaced vector
where each interval is equally distributed from $-\pi$ to $\pi$.
In this scheme, it is crucial to properly determine the frequency vector
$\boldsymbol{\omega}:D_{\text{p}}\times1$, in which the highest
value referred to maximum frequency $\omega_{\text{max}}=(N_{\text{se}} -1)/2M_{\text{e}}$
is assigned to the parameter under study.
$M_{\text{e}}$ is known as the interference factor
and acts as the remover for numerical amplitude from superposing of waves.
From the empirical investigation \citep{Saltelli2002,Marino2009},
it is usually used as 4 or 6 (we use 4 in this work).
Other frequencies in $\boldsymbol{\omega}$ are set in the
range $[1,\ \omega_{-i,\text{max}}]$ with a regular increment
$\frac{\omega_{-i,\text{max}}}{D_{\text{p}}}$, where
$\omega_{-i,\text{max}}=\frac{\omega_{\text{max}}}{2M_{\text{e}}}$.

\begin{figure}[ht]
\centering
\includegraphics[width=0.40\textwidth]{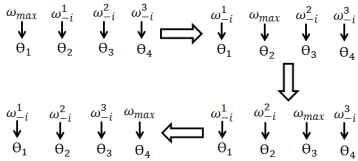}
\caption{
When we wise to evaluate the sensitivity of parameter $\theta_{1}$,
its samples are drawn with the highest frequency $\omega_{\text{max}}$, while
the samples for other parameters $\boldsymbol{\theta}_{-i}=\{\theta_{2},\ \theta_{3},\ \theta_{4}\}$
in the system are picked using
the complementary frequencies $\boldsymbol{\omega}_{-i}=\{\omega^{1}_{-i},\ \omega^{2}_{-i},\ \omega^{3}_{-i}\}$.
Through this process, all parameters
in system should be assigned to the highest frequency once.
}
\label{Figure:FrequencyAssign}
\end{figure}

This strategy means that the sensitivity of underlying parameter
is assessed by picking the samples with the highest frequency $\omega_{\text{max}}$,
while the samples for the rest of the parameters are selected with the complementary frequencies
$\boldsymbol{\omega}_{-i}$.
This process is repeated until the samples of each parameter is
drawn with highest frequency once.
An illustrative example of this cycling process is shown in
Fig. \ref{Figure:FrequencyAssign}.

Consequently, the parameter can be seen as the most sensitive when which has the largest normalized variance, i.e. its variation causes the highest uncertainty.
In addition, for any sample of parameter drawn from the unit hypercube $\boldsymbol{K}^{i}= (\boldsymbol{\theta}|0\leq \theta_{i}\leq 1;\ i=1,\ldots,D_{\text{p}})$, it has to be re-scaled to real-valued space as such the model evaluation can be carried out. Hence, a non-informative prior covered whole range of parameter value is required.
In this work, we consider the suggestion from \cite{Marino2008}, where authors claimed that eFAST artificially yields small but non-zero sensitivity indexes for parameters, and thus a `dummy' parameter which does not really exist in model and has no influence on behaving dynamics in any way, is introduced in eFAST to absorb such artifacts.
The complete description of eFAST including the mathematical derivation and pseudo-code are presented in \textbf{Section 3} of \textbf{Supplementary materials}.

\color{black}

\section{Results and Discussion}
We demonstrate the effectiveness of the proposed method using three models of sustained periodic oscillations and and one model of transient response to shock. These are models used by previous authors, including ourselves, enabling comparisons to be made. We give details of implementation in \textbf{Section 6} of \textbf{Supplementary materials}. Crucial results from the heat shock and repressilator systems are discussed here and results from the remaining two systems are given in \textbf{Supplementary materials}.

\begin{figure*}[ht]
\centering
\includegraphics[width=0.85\textwidth]{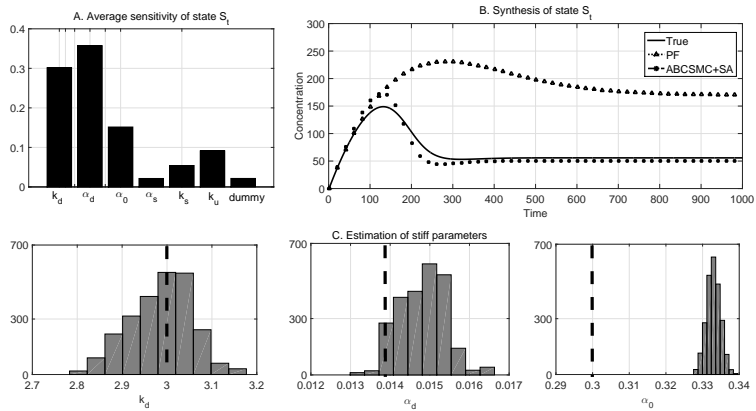}
\caption{
Sensitivity analysis, inference of parameters and
system re-synthesis of heat shock model.
A: Average sensitivity of parameters with respect to
state $S_{t}$. B: Reproduction of state $S_{t}$ by using true values,
estimates from ABC+SA (dotted line) and PF (dotted triangle) respectively.
C:Histograms for the
stiff parameters $k_{d}$, $\alpha_{d}$ and $\alpha_{0}$.
The black lines indicate the true values of parameters proposed in
the literature.}
\label{Figure:ABCSA_HS}
\end{figure*}

\textbf{The heat shock model}
We consider the case of estimating all six parameters of this model simultaneously. This is the most difficult case considered in our previous work \citep{Xin2012}, in which particle filters were able to correctly estimate four of the six unknown parameters, and failed in the remaining two ($k_d$ and $\alpha_d$). Fig. \ref{Figure:ABCSA_HS}.A shows results of sensitivity analysis of one of the states $S_t$ (see equations of the system in \textbf{Supplementary materials}). Of these, parameters $\alpha_{d}$, $k_{d}$ and $\alpha_{0}$, to which $S_t$ is most sensitive to, are treated as stiff and the remainder as sloppy. Posterior distributions over estimates of the stiff parameters, using the proposed method, are shown in Fig. \ref{Figure:ABCSA_HS}.C and the corresponding sloppy ones are given in \textbf{Fig. S1(c)}. While the posterior means of $k_d$ and $\alpha_d$ are estimated to a high level of accuracy, $\alpha_0$ is not. This is explained by the sensitivity results in which we note that, though we took all these three to be stiff parameters, the state is not as sensitive to $\alpha_0$ as it is to $k_d$ and $\alpha_d$. Hence, relatively poor estimates of $\alpha_0$ will still lead to accurate reconstruction of the model outputs.

In Fig. \ref{Figure:ABCSA_HS}.B, we show two responses synthesized from the posterior mean values of particle distributions from a classic particle filtering approach (as considered in our previous work, \cite{Xin2012}) and the ABC-SMC method proposed here. We note that in this difficult problem of estimating all six parameters of the model, ABC-SMC is able to successfully find solutions of parameters from which the underlying state could be accurately generated for the state $S_{t}$ in the heat shock model. We observe the same to be true for the other two states, $D_{t}$ and $U_{f}$ as well (graphs shown in \textbf{Fig. S1.(a)} of \textbf{Supplementary materials}).

\textbf{Repressilator system}
The deterministic repressilator system is constructed as a synthetic gene regulatory
circuit which can sustain oscillations by the mutual repression of gene transcription \citep{Elowitz2000}.
This system consists of six differential equations, from three pairs of mRNA and protein,
and has four parameters. This system was analysed by \cite{Tina2009} to demonstrate
their ABC-SIS approach.

\begin{figure*}[ht]
\centering
\includegraphics[width=0.85\textwidth]{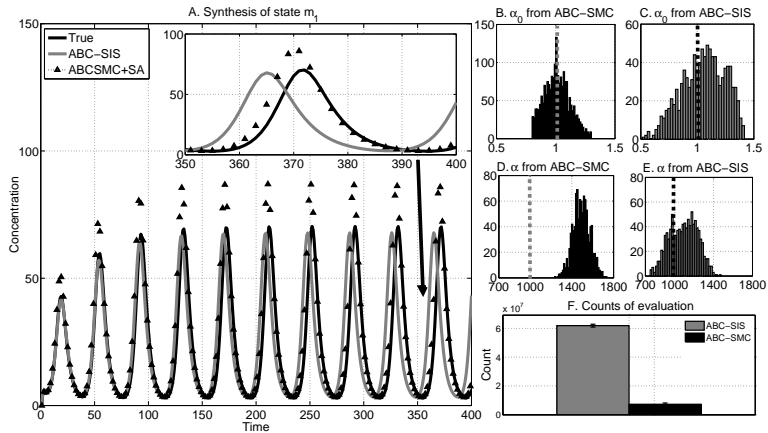}
\caption{
System synthesis, inference and computational performance:
A: Synthesis of system behavior by using the true values from literature,
inferred values from two methods. The proposed method (denoted by dotted triangle)
captures the system behavior in terms of periodicity, but with larger amplitude.
ABC-SIS successfully recovers the maximum, while a slight offset occurs in the last
few time instants. The zoom-in version concentrates the last periodicity.
The clear and color version of these graphs are given as \textbf{Fig.S2}.
B-C: The estimates of stiff parameter $\alpha_{0}$ from ABC-SMC+SA and ABC-SIS.
D-E: The estimates of sloppy parameter $\alpha$ from ABC-SMC+SA and ABC-SIS.
In these graphs, the dash lines indicate the true values of corresponding
parameters.
F: Counts of model evaluation taken by ABC-SIS
and the ABC-SMC+SA to achieve the final tolerance $\epsilon_{T}$.
}
\label{Figure:ABCSA_Repressilator}
\end{figure*}

While we have set the posterior means of a coarse search to sloppy parameters ($\alpha$ and $\beta$), ABC-SMC is capable to find accurate solutions of stiff parameters, given allocated extra computational budget.
Posterior distributions of a stiff ($\alpha_0$) and sloppy ($\alpha$) parameters from our proposed method and its predecessor ABC-SIS are shown in Fig.\ref{Figure:ABCSA_Repressilator}.B-E. We note that, with an identical tight acceptance condition $\epsilon_{T}$, both methods are adequate in estimating stiff parameters to a relatively high level of accuracy. However, ABC-SIS shows a distinguishable advantage in identifying sloppy parameters, whereas our proposed method fails to achieve convergence to the correct estimate. Reconstruction of model behavior, using the posterior mean values from both methods, tells that the failure of finding solutions for sloppy parameters does not hugely hamper the effectiveness in characterizing dynamics. The complete results are shown as color figure in $\textbf{Fig. S2-S3}$ of $\textbf{Supplementary materials}$. In addition, we also tested how sensitivity analysis affects ABC-SMC by setting all possible combinations of stiffness/sloppiness to parameters, we found that at lease one stiff parameter is correctly chosen, ABC-SMC is able to find solutions. Results are given in \textbf{Table S7}.

In order to show how ABC-SIS is influenced by the suboptimal user-specific algorithmic setting, we compare the computational expense between ABC-SIS and our proposed method using identical initial and final tolerances. Since ABC-SMC on average requires 140 iterations to reach $\epsilon_{T}$, we therefore reduce the tolerance from $\epsilon_{0}$ down to $\epsilon_{T}$ using 140 steps at regular intervals. The algorithm efficiency is quantified by counting the number of model evaluations, at which most of computational budget is spent, to reach final tolerance $\epsilon_{T}$.
Fig. \ref{Figure:ABCSA_Repressilator}.F clearly shows that the number of evaluations made by ABC-SIS is approximately
seven times greater than ABC-SMC.
This is because the suboptimal chosen transition kernel
and tolerance schedule make ABC-SIS expend considerable
computation on searching for acceptable particles.
\color{black}

\textbf{Other systems}
We further consider delay-driven oscillatory system
and cell cycle system, for which the complete descriptions
and results are given in \textbf{Section 4.3} and
\textbf{Section 4.4} of \textbf{Supplementary materials}.

In the delay-driven oscillatory example,
for testing the general advantage of proposed method,
we use the non-sophisticated ABC-MCMC \citep{Marjoram2003}
instead of ABC-SMC in this example.
From the results given in \textbf{Fig. S4},
it is easy to observe that the original ABC-MCMC poorly performs with the
coarse acceptance criterion,
while it can produce the accurate inference by using
a small tolerance $\epsilon$.
When the similar precision of inferences
is achieve, as expected, the proposed method
greatly outperforms in efficiency
(the comparison is shown in \textbf{Fig. S4.F}).

The cell cycle system, with 12 free parameters in the model, is a relatively high dimensional system. When particles are sampled in this space, we found that for a significant fraction of the samples, the ODE solver failed to terminate. With the proposed method, because sensitivity analysis identifies the sloppy parameters which have been set to the posterior means of a coarse search, the dimensionality is reduced and this issue gets circumvented.

However, since the state $RA$ in this cell cycle
has no influential parameter in behaving dynamics (sensitivity analysis
with respect to $RA$ is given in \textbf{Fig. S8}),
therefore, a favorable computational budget allocation
for state $RA$ can only be provided by associating
the higher-order sensitivity index (influence on outputs
caused by the interaction between parameters).

\begin{figure*}[ht]
  \centering
  \includegraphics[width=0.85\textwidth]{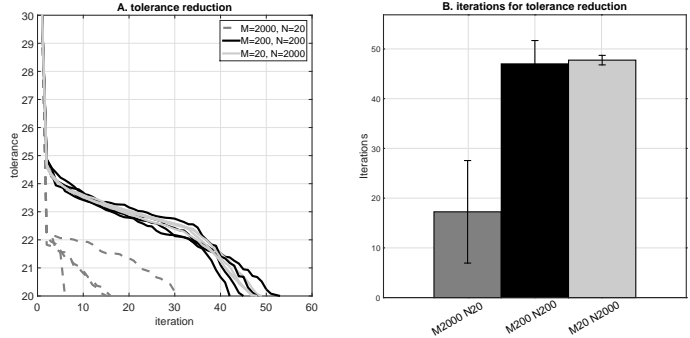}
  \caption{Effect of $M_{\text{smc}}$ versus $N_{\text{smc}}$ on approaching target tolerance $\epsilon_{T}$ . The total amount of particles remains the same, i.e. 40,000, and particles are distributed in different settings of $M_{\text{smc}}$ and $N_{\text{smc}}$. The above example, three combinatins are examined: $M_{\text{smc}}$ = 2, 000 and $N_{\text{smc}}$ = 20; $M_{\text{smc}}$ = 200 and $N_{\text{smc}}$ = 200; $M_{\text{smc}}$ = 20 and $N_{\text{smc}}$ = 2, 000. Each combination is carried out 4 times on the repressilator system. A: Trajectories of approaching target tolerance $\epsilon_{T}$ by different combinations. B: Average iterations carried out by different combinations. Overlapping graphs are presented more clearly as color figure in \textbf{Fig. S12} of \textbf{Supplementary materials}.}
  \label{comparison}
\end{figure*}

\textbf{Effect of $M_{\text{smc}}$ and $N_{\text{smc}}$}
In running ABC-SMC algorithms, we would ideally like to
choose large values for $M_{\text{smc}}$ and $N_{\text{smc}}$ for achieving high accuracy
of exploring the posterior density. Naturally, this will increase
the computational complexity of the implementation. How does
one trade-off spending a given computational budget across $M_{\text{smc}}$
(the number of parallel filters) and $N_{\text{smc}}$ (the number of particles
per filter)? The choice is likely to be problem dependent, but an
empirical exploration, using the repressillator system, over three
sets of values for these $M_{\text{smc}}$ = 20 and $N_{\text{smc}}$ = 2, 000; $M_{\text{smc}}$ =
200 and $N_{\text{smc}}$ = 200; and $M_{\text{smc}}$ = 2, 000 and $N_{\text{smc}}$ = 20, keeping
their product at 40, 000, is shown in Fig. \ref{comparison}. The corresponding
accuracies of inference are given in Table \ref{ABCSA_RRMSE_M_N}.

\begin{table}[h]
\caption{Comparison of NRMSE for different settings of $M_{\text{smc}}/N_{\text{smc}}$}
\label{ABCSA_RRMSE_M_N}
\centering
\begin{tabular}{|c|c|c|c|c|c|}
\hline
\multicolumn{2}{|l|}{} & \multicolumn{4}{|c|}{NRMSE}  \\
\hline
$M_{\text{smc}}$          & $N_{\text{smc}}$         &  $\alpha_0$ & $n$   & $\beta$ & $\alpha$ \\ 
2000       & 20        &  6.41$\pm$2.71   & 1.92$\pm$0.20 & 1.01$\pm$0.20  & 1.04$\pm$0.34   \\
200        & 200       &  6.17$\pm$0.72   & 1.21$\pm$0.04 & 1.21$\pm$0.10  & 1.11$\pm$0.10   \\
20         & 2000      &  5.20$\pm$0.45  & 1.11$\pm$0.04 & 1.03$\pm$0.05  & 0.99$\pm$0.02   \\
\hline
\end{tabular}
\end{table}

As seen in the graphs, smaller values of  $M_{\text{smc}}$ are computationally less expensive.
\color{black}This is due to lower value of $M_{\text{smc}}$ causing a higher
probability of particles being `killed' and having a zero weight.
As as consequence, decrements taken in $\boldsymbol{\epsilon}$ are smaller
slowing convergence to its target $\epsilon_{\text{T}}$.
For instance, considering the large $M_{\text{smc}}$
example, i.e. $M_{\text{smc}}$ = 2000, evaluating importance using Eqn.\ref{weightABCSMC},
the zero weight barely appears. Intuitively, the greater non-zero
proportion of the weight vector implies that most particles fulfill
the current acceptance criterion, therefore, a large decrement should
be taken for the next tolerance level.
From the perspective of accuracy, when gaining a larger number
of particles, the diversity of realizations increases and better
performance in terms of precision is naturally expected. This
conclusion is verified by the comparison of Normalized Root
Mean Squared Error (NRMSE) which is a measurement to
quantify the quality of prediction and its expression is given as:
\begin{align}\label{NRMSE}
\text{NRMSE} = \frac{\sqrt{\frac{\sum_{i=1}^{N}{\widehat{\theta}_{i}-\theta_{\text{true}}}}{N}}}{\max({\widehat{\boldsymbol{\theta}}})-\min({\widehat{\boldsymbol{\theta}}})}
\end{align}
From the results shown in Table \ref{ABCSA_RRMSE_M_N}, the higher $N_{\text{smc}}$
value generally delivers the better inference, where $N_{\text{smc}}$ = 2000
performs best among all considered settings.

\textbf{Why use eFAST for sensitivity analysis?}\label{whyEFAST}
We have chosen eFAST, a global method, for sensitivity analysis in this work. The search carried out by this method, systematically exploring a wide range in the parameter space is computational expensive. An alternate method of quantifying parameter sensitivities, particularly in the context of particle filtering, adopted by \cite{Tina2009} is based on principal component analysis (PCA) of the particle distribution, and to use the principal directions as indicators of sloppy and stiff directions of parameter combinations. Parameters that are well aligned to the directions may show the declared \color{black} stiff and sloppy. However, this simplistic approach is inadequate for our purpose for two reasons. Firstly, we wish to identify sloppy and stiff parameters of the models before running the particle filters, hence PCA on the distribution of a converged solution is not very useful. Secondly, and somewhat more importantly, PCA is a good representation of data that is multivariate Gaussian. If we expect the distribution of particles to be multivariate Gaussian, we would not be running particle filtering anyway, because Kalman filter type updates of posterior means and covariances are the known optimal solutions. In order to confirm this, we took the distributions of particles, after convergence, of all the four models considered and tested them for multivariate Gaussianity using  Henze-Zirkler (multivariate normality) test \citep{Henze1990}. Such test is carried out upon measuring the non-negative functional distance between two distributions. If data of interest are distributed as multivariate normal, the test statistic is approximately log-normally distributed. The function {\tt hzTest} from R package {\tt MVN} \citep{Korkmaz2014} is used in this work\color{black}.
From the test, we found three of them to be non-Gaussian. This confirms that the choice of global variance based sensitivity analysis is the correct one for our analysis.

\section{Conclusion}
In this work, we proposed an inference method for analyzing Systems Biology models that couples sensitivity analysis and approximate Bayesian computation. Our proposed method is particularly advantageous in difficult settings of estimating all (or most of) the parameters of a model from noisy observations, because it strikes a balance between accuracy and efficiency. The method exploits the fact that all parameters in model have different significance in characterizing model dynamics in terms of their sensitivities. By re-synthesis data from models with estimated parameters, we show that the values of parameters that are more critical (stiff parameters) need to be determined with care, while the sloppy parameters need not be estimated to high precision. To facilitate such inference, we have proposed a three stage strategy in which a selective computational budget allocation is implemented via sensitivity analysis, in which the sloppy group is estimated by a coarse search followed the re-estimation of stiff parameters to tighter error tolerances.

We have demonstrated the effectiveness of the proposed approach on
three systems of oscillatory behavior and one of transient response.
The results show that the introduction of favorable inference strategy
allows to reduce the dimension of unknown parameters,
and paves a potential way to tackle the complex problems.
Additionally, the used ABC-SMC has attracted much interest
due to its adaptivity in determining tolerance schedule $\boldsymbol{\epsilon}$
and the `no rejection' of particles allows to boost the efficiency
via parallel computing.
In the simple problem, e.g. the delay-driven oscillatory system,
with performing similarly in accuracy,
the proposed inference method expends much less computational cost
than the existed ABC methods.           

\bibliography{ECS}
\end{document}